\documentclass{article}

\usepackage{PRIMEarxiv}

\usepackage[utf8]{inputenc} 
\usepackage[T1]{fontenc}    
\usepackage{hyperref}       
\usepackage{url}            
\usepackage{booktabs}       
\usepackage{amsfonts}       
\usepackage{nicefrac}       
\usepackage{microtype}      
\usepackage{lipsum}
\usepackage{fancyhdr}       
\usepackage{graphicx}       
\graphicspath{{media/}}     

\usepackage{colortbl}
\usepackage{algorithm}
\usepackage{algorithmic}
\usepackage{xcolor}
\usepackage[authoryear,square]{natbib}
\let\cite\citep
\usepackage{amsmath}

\title{EvoPrune: Early-Stage Visual Token Pruning for Efficient MLLMs
}

\author{
  Yuhao Chen\thanks{Work done during an internship at ByteDance.} \\
  chenyuhao.101@bytedance.com \\
  ByteDance \\
  Beijing, China\\
   \And
  Bin Shan \\
  shanbin@bytedance.com \\
  ByteDance \\
  Beijing, China\\
   \And
  Xin Ye \\
  yexin.93@bytedance.com \\
  ByteDance \\
  Beijing, China\\
   \And
  Cheng Chen \\
  chencheng.kit@bytedance.com \\
  ByteDance \\
  Beijing, China\\
}

\begin{document}
\maketitle

\begin{abstract}
Multimodal Large Language Models (MLLMs) have shown strong performance in vision-language tasks, but their inference efficiency is severely limited by the exponential growth of visual tokens in complex scenarios such as high-resolution images and videos. Existing visual token pruning methods mainly operate after visual encoding, overlooking the substantial computational cost incurred during the encoding stage.
To address this issue, we propose EvoPrune, an early-stage visual token pruning method for MLLMs that performs pruning directly during visual encoding. Specifically, EvoPrune employs a layer-wise pruning strategy guided by token similarity, diversity, and attention-based importance to retain the most informative visual tokens at selected encoding layers.
Extensive experiments on image and video benchmarks validate the effectiveness of EvoPrune. In particular, on the VideoMME dataset, EvoPrune achieves 2$\times$ inference speedup with less than 1\% performance degradation, demonstrating its potential for latency-sensitive MLLM deployment.
\end{abstract}
\section{Introduction}
\label{sec:intro}
\begin{figure}[htbp!]
  \centering
  \includegraphics[width=0.7\textwidth]{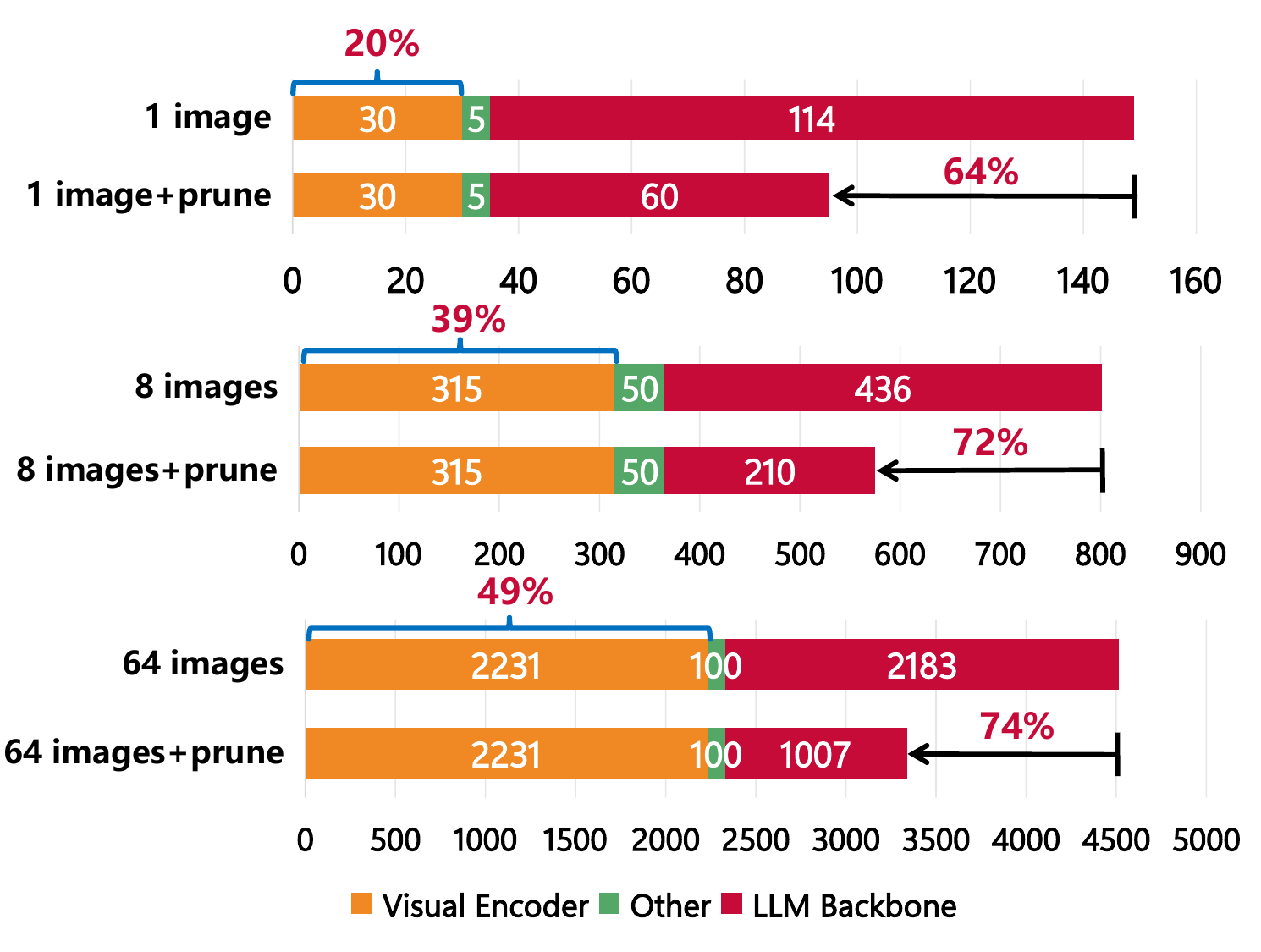}
  \caption{\textbf{Inference time of different components in MLLMs with and without visual Pruning under different input scales.}}
  \label{fig:first}
\end{figure}

Multimodal Large Language Models (MLLMs) \cite{wang2024qwen2,li2024llava,liu2023visual,bai2025qwen2,achiam2023gpt,team2024gemini,chen2024internvl,lu2024deepseekvl,guo2025seed1,liu2024improved} have demonstrated substantial progress across diverse vision-language tasks, such as visual question answering, thereby greatly broadening their practical applicability. However, in high-resolution image processing and long-sequence video understanding scenarios, the number of visual tokens increases dramatically, resulting in severe inference inefficiencies. Specifically, the explosion of visual tokens incurs significant computational and memory overhead in both the visual encoding and LLM prefill stages, forming a critical efficiency bottleneck that limits the deployment of MLLMs in latency-sensitive settings, including real-time video analysis and edge computing.

To address visual token explosion, existing approaches can be broadly categorized into two paradigms. The first paradigm \cite{li2024llava,dai2023instructblip,li2023blip} relies on learned projectors trained on large-scale datasets for token selection, achieving moderate dimensionality reduction but suffering from limited generalization and the need for retraining. The second paradigm performs direct token pruning during inference based on visual token representations. Within this category, attention-based methods \cite{chen2024image,zhang2024sparsevlm,xing2024pyramiddrop,yang2025visionzip} retain salient tokens using attention scores, while similarity-based approaches \cite{yang2025visionzip,alvar2025divprune,zhang2025beyond,shang2025llava} eliminate redundancy via unsupervised similarity analysis (e.g., determinantal point processes \cite{chen2018fast}). These methods demonstrate notable inference acceleration with minimal performance degradation.

Despite their effectiveness, existing vision-token pruning methods share a fundamental limitation: pruning is applied only after visual encoding. As shown in Fig.~\ref{fig:first}, the computational cost of the vision encoder increases rapidly as the input scale grows from a single image to multi-frame inputs (8 and 64 frames), eventually approaching the inference cost of the LLM backbone. However, most prior methods perform pruning at intermediate stages or within the LLM, primarily reducing language-model computation while leaving the dominant encoder cost unchanged. Consequently, pruning effectiveness degrades with increasing input scale. As illustrated in Fig.~\ref{fig:first}, the remaining inference time rises from 64\% (single image) to 72\% (8 frames) and 74\% (64 frames) of the original runtime, indicating that existing approaches yield diminishing acceleration and encounter a scalability bottleneck under large visual inputs.

To overcome these limitations, we introduce \textbf{EvoPrune}, an early-stage vision-token pruning framework for MLLMs. EvoPrune removes redundant tokens at the initial stages of visual encoding, prior to expensive feature computation, thereby substantially reducing the computational cost of both the vision encoder and the downstream LLM. This design enables significant end-to-end inference acceleration across varying input resolutions.

Unlike existing methods that apply token pruning only after full visual feature extraction, \textbf{EvoPrune} integrates pruning directly within the visual encoder. At selected layers, it progressively merges redundant or low-importance tokens using a multi-criteria strategy that leverages token similarity, diversity, and intermediate attention signals. This hierarchical, guided pruning process preserves salient visual information while markedly reducing the computational overhead of the vision encoder and subsequent multimodal components, facilitating efficient inference across diverse visual inputs.

We conduct evaluations on multiple benchmarks spanning image and video understanding tasks. Results show that EvoPrune achieves state-of-the-art efficiency-performance trade-offs. In particular, on the challenging VideoMME benchmark, EvoPrune reduces inference latency by 50\% with less than a 1\% performance degradation. Overall, our contributions can be summarized as follows:
\begin{enumerate}
    \item We propose a novel early-stage pruning paradigm, which acts in the visual encoding stage to address the long-neglected encoding overhead in existing works.
    \item We adopt a layer-wise pruning approach with multi-factor guidance, integrating token similarity, diversity, and attention-derived importance to guide token selection and maximal information retention at each layer.
    \item We validate EvoPrune across diverse vision-language tasks, demonstrating that it outperforms existing methods in inference efficiency while retaining competitive task performance.

\end{enumerate}

\section{Related work}

\subsection{Multimodal Large Language models}
Large-scale multimodal models \cite{wang2024qwen2,li2024llava,lu2024deepseekvl,liu2023visual,bai2025qwen2,achiam2023gpt,team2024gemini,wang2025pargo,li2023blip,liu2024improved} have achieved remarkable progress, facilitating their expansion to more complex tasks. These models feed visual-encoder representations into Large Language Models (LLMs) after specific processing (e.g., via a projector). With the growing prevalence of applications, community researchers have found that real-world scenarios frequently demand the comprehension of large-scale images, videos, or sequential image frames. 

Recently, several works \cite{liu2024llavanext,liu2024textmonkey,shi2025scaling,zhang2024llavauhdv2,zhang2024beyond} have focused on improving the performance of diverse tasks with large-scale image inputs. Meanwhile,  another stream of  works \cite{zhang2024llavanext-video,chen2024longvila,shen2024longvu,shu2025video,wang2025internvideo2} have extended MLLMs to the field of large-scale video understanding. Nevertheless, as task scenario complexity increases, the number of visual tokens escalates substantially in specific tasks—such as high-resolution image recognition or video understanding. Therefore, mitigating the model’s computational overhead resulting from the significant surge in visual tokens under these scenarios has become a critical challenge.

\subsection{Visual Token Pruning}
Visual token pruning has been extensively studied to reduce the quadratic cost in vision transformers. Early works~\cite{bolya2022token,rao2021dynamicvit,li2024vidtome} primarily focus on standalone visual models, and explore token pruning for image classification and recognition tasks, aiming to remove spatial redundancy within ViTs.

With the rise of MLLMs, recent studies extend visual token pruning to long-context settings, including high-resolution images and long videos. Unlike pure vision tasks, MLLMs must preserve fine-grained semantics and cross-modal alignment.
Existing methods\cite{wang2024qwen2,guo2025seed1,li2024llama,wang2025pargo,cha2024honeybee} fall into four categories by core mechanism: transformation-based methods use multi-scale feature fusion for compact representation; similarity-based methods\cite{yang2025visionzip,alvar2025divprune,zhang2025beyond,shen2024longvu,shao2025holitom,song2024moviechat,zhang2024cls} merge semantically/spatially adjacent tokens to reduce redundancy; attention-driven methods \cite{shang2025llava,zhang2025attention,zhang2025vscan,zhang2024sparsevlm,zhang2025beyond} select key tokens via weight analysis; query-based methods\cite{liu2024multi,wang2025pargo,li2023blip,dai2023instructblip,song2025less} perform goal-oriented pruning with task-specific queries.
Video-specific pruning\cite{shao2025holitom,chen2018fast,shen2024longvu,song2024moviechat,shen2025fastvid} adds temporal strategies (e.g., key frame sampling, cross-frame merging) to effectively reduce inter-frame redundancy. 
Current methods’ core challenge is balancing compression efficiency, information retention, and multimodal compatibility. These works lay a foundation for visual token optimization in long-context multimodal processing and support MLLMs’ efficient deployment.
\section{Method}
\subsection{Overall}
Visual token pruning is an effective approach to reducing the computational cost of MLLMs. However, most existing methods apply pruning only after the visual encoder processes all visual tokens, resulting in limited efficiency gains and a pronounced computational bottleneck, especially for high-resolution images and long videos.

To overcome this limitation, we propose \textbf{EvoPrune}, an \emph{early-stage} visual token pruning framework that integrates progressive token merging directly into the visual encoder. By merging low-importance or redundant tokens across multiple encoder layers, EvoPrune enables early-stage acceleration throughout multimodal inference while preserving task-critical visual information.

\begin{figure}[t]
  \centering
  \includegraphics[width=0.6\linewidth]{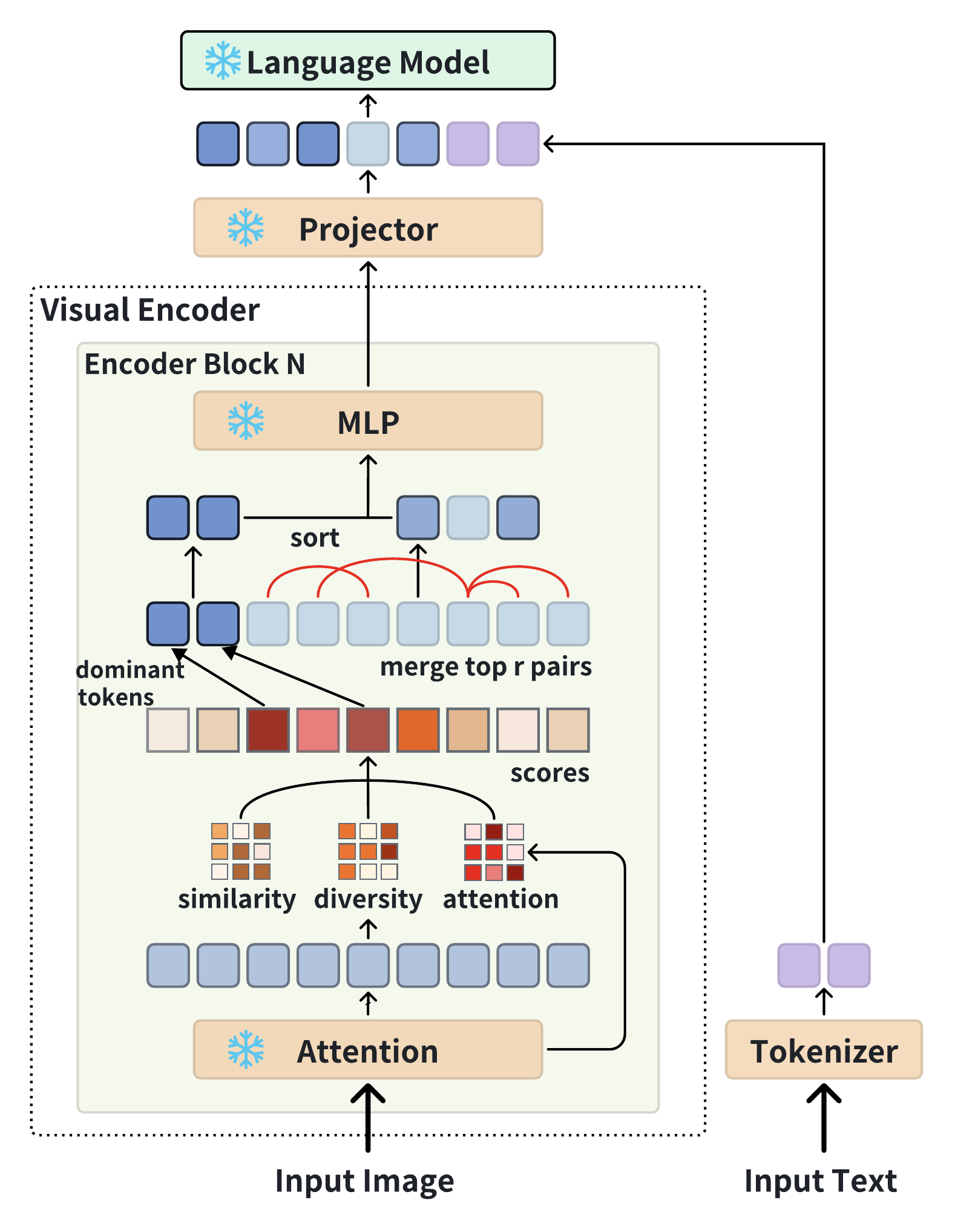}
  \caption{\textbf{Overview of the EvoPrune framework.} Token merging operations are applied at selected visual encoder layers. A composite score matrix integrating semantic similarity, diversity, and attention-based importance, guides token pair selection.
  }
  \label{fig:architecture_overall}
\end{figure}

Figure~\ref{fig:architecture_overall} presents an overview of the EvoPrune framework. At selected visual encoder layers, tokens are first processed by standard multi-head self-attention. Token merging is then performed based on a composite \emph{score matrix} that integrates three complementary criteria:
\begin{itemize}
  \item \textbf{Semantic Similarity}, which promotes merging of visually and semantically redundant tokens;
  \item \textbf{Information Diversity}, which discourages merging tokens carrying distinct content to maintain representational richness;
  \item \textbf{Attention-Based Importance}, which preserves tokens critical to downstream reasoning.
\end{itemize}
Token pairs with the highest composite scores are merged and propagated to subsequent layers. This hierarchical pruning process progressively reduces the token set while retaining discriminative visual representations.

By embedding pruning into the visual encoder, EvoPrune alleviates the computational bottleneck of post-encoder pruning and achieves efficient multimodal inference without sacrificing visual fidelity.

\subsection{Layer-wise Pruning Budget Allocation}
EvoPrune adopts a layer-wise pruning formulation that explicitly allocates token merging budgets across selected visual encoder layers.

\begin{algorithm}[htbp]
\caption{Layer-wise Token Pruning in EvoPrune}
\label{alg:layerwise_pruning}
\begin{algorithmic}[1]
\REQUIRE Initial visual tokens $\mathcal{T}_0$;
         visual encoder with $L$ transformer blocks;
         total pruning budget $R$;
         layer-wise merging strategy $\mathcal{S}$
\ENSURE Encoded visual tokens $\mathcal{T}_L$

\STATE Allocate the total pruning budget $R$ across encoder layers using strategy $\mathcal{S}$,
       yielding a layer-wise schedule $\mathbf{r} = \{r_1, r_2, \dots, r_L\}$,
       where $\sum_{l=1}^{L} r_l = R$

\STATE $\mathcal{T} \leftarrow \mathcal{T}_0$

\FOR{$l = 1$ to $L$}
    \STATE \textit{// Multi-head self-attention}
    \STATE $\mathcal{T} \leftarrow \text{Attention}_l(\mathcal{T})$

    \STATE \textit{// In-encoder token pruning}
    \IF{$r_l > 0$}
        \STATE $\mathcal{T} \leftarrow \text{ScoreGuidedMerge}(\mathcal{T}, r_l)$
    \ENDIF

    \STATE \textit{// Feed-forward network}
    \STATE $\mathcal{T} \leftarrow \text{MLP}_l(\mathcal{T})$
\ENDFOR

\STATE \textbf{Return} $\mathcal{T}$
\end{algorithmic}
\end{algorithm}

Algorithm~\ref{alg:layerwise_pruning} outlines the pruning process. Given input images or videos, a global pruning target, and a predefined merging strategy, EvoPrune determines how many tokens to merge at each encoder layer during the forward pass.

Formally, consider a visual encoder with $L$ transformer layers. Given an initial token set $\mathcal{T}_0$ containing $N_0$ tokens and a global pruning target $R$, EvoPrune computes a layer-wise pruning schedule
\[
\mathbf{r} = \{r_1, r_2, \dots, r_L\}, \quad \text{s.t. } \sum_{l=1}^{L} r_l = R,
\]
where $r_l \geq 0$ denotes the number of tokens merged at the $l$-th layer.

During inference, token merging is applied at layer $l$ if $r_l > 0$ using the score-guided mechanism described in Section~\ref{sec:Score-Guided Token Merging}.

\paragraph{\textbf{Merging Strategy Design}}
Due to space limitations, we defer the detailed design of layer-wise merging strategies to the appendix. We systematically explore multiple budget allocation strategies and analyze their effects on efficiency and accuracy. Full formulations and experimental results are provided in Appendix \ref{sec:appendix_merging_strategy}.

\subsection{Score-Guided Token Merging}
\label{sec:Score-Guided Token Merging}
We propose a score-guided token merging strategy specifically designed for in-encoder pruning in MLLMs. Our method introduces a \emph{score matrix} that evaluates the potential benefit of merging token pairs using multiple complementary factors, making it suitable for the multimodal and information-sensitive setting of MLLMs.

Specifically, we first partition the token set 
\[
\mathcal{T} = \{t_1, t_2, \dots, t_N\}
\] 
into two disjoint groups \(a\) and \(b\) according to their indices:
\[
a = \{t_i \in \mathcal{T} \mid i \bmod 2 = 0\}, \quad 
b = \{t_i \in \mathcal{T} \mid i \bmod 2 = 1\}.
\]
Given these groups, we define a score matrix
\[
Score \in \mathbb{R}^{|a| \times |b|},
\]
where each entry \(Score_{ij}\) measures the potential gain of merging token \(t_i \in a\) into token \(t_j \in b\). The matrix is computed exclusively for cross-group pairs.

To determine merge candidates, for each token \(t_i \in a\), we identify the token in \(b\) with the highest score:
\[
j^* = \displaystyle \arg\max_{j \in b} Score_{ij}.
\]
This ensures that each token in \(a\) participates in at most one merge. Among the resulting candidate pairs 
\(\{(i,j^*) \mid t_i \in a\}\), the top-\(r\) pairs with the highest scores are selected:
\[
\mathcal{M} = \text{Top-}r \bigl\{(i,j^*) \mid Score_{ij^*} \bigr\}.
\]
Finally, the selected merges are applied from group \(a\) to \(b\), and the remaining tokens are reorganized to form the updated token set \(\mathcal{T}'\) for the next processing stage.

This strategy establishes a principled and flexible framework for visual token merging, in which each merge operation is explicitly defined and controlled, while preserving the structural consistency of the token set for downstream processing. The computation and formulation of the score matrix \(Score\) are described in the subsequent subsection.

\subsection{Similarity, Diversity, and Attention Integration}
To guide the token merging process within the visual encoder, the overall matching score between tokens \( i \) and \( j \) is formulated as:
\[
\text{Score}_{ij} =
\underbrace{\text{Sim}_{ij}}_{\text{Similarity Attraction}} \times
\underbrace{w_{ij}^{\text{div}}}_{\text{Diversity Penalty}} \times
\underbrace{w_{ij}^{\text{attn}}}_{\text{Attention Preservation}}.
\]
where:
\begin{itemize}
  \item \( \text{Sim}_{ij} \): the semantic similarity between visual tokens \( i \) and \( j \),
  \item \( w_{ij}^{\text{div}} \): the redundancy to encourage token diversity,
  \item \( w_{ij}^{\text{attn}} \): the attention distribution learned by the encoder.
\end{itemize}
The composite score comprises three complementary terms: \textbf{Similarity Attraction}, \textbf{Diversity Penalty}, and \textbf{Attention Preservation}, jointly balancing token compactness and information retention. Based on this score, we perform hierarchical token merging across encoder layers using a bipartite soft-matching scheme that iteratively aggregates token pairs.

In contrast to prior token merging methods driven solely by pairwise similarity, our formulation explicitly incorporates diversity and attention constraints, yielding more discriminative and information-preserving visual representations.

\paragraph{\textbf{Similarity Attraction}}
The similarity attraction component quantifies the visual correspondence between tokens \( i \) and \( j \) based on the cosine similarity of their feature embeddings \( a_i \) and \( b_j \):
\[
\text{Sim}_{ij} = \frac{a_i \cdot b_j}{\|a_i\| \, \|b_j\|}.
\]
\noindent
The similarity score \( \text{Sim}_{ij} \) ranges within \( [0, 1] \) and reflects the angular proximity between token representations in the embedding space. 
For improved numerical stability and comparability across tokens, the similarity values can be optionally normalized using a sigmoid or softmax function before being integrated into the overall matching score.

\paragraph{\textbf{Diversity Penalty}}

To encourage structural diversity during token merging, we evaluate the local distinctiveness of each token embedding \( \mathbf{v}_i \in \mathbb{R}^d \) by estimating its local density. The local density \( d_i \) quantifies the degree of uniqueness of token \( i \) relative to its neighborhood, where \( \mathcal{N}_i \) denotes the set of its \( K \) nearest neighbors determined by the squared Euclidean distance.
\[
d_i = 1 - \frac{1}{K} \sum_{k \in \mathcal{N}_i} 
\exp\!\left(-\frac{\| \mathbf{v}_i - \mathbf{v}_k \|^2}{\tau^2}\right).
\]
\noindent
where:
\begin{itemize}
  \item \( \mathcal{N}_i \) is the set of the \( K \) nearest neighbors of \( \mathbf{v}_i \),
  \item \( \tau > 0 \) denotes the bandwidth parameter of the Gaussian kernel,
  \item \( \| \mathbf{v}_i - \mathbf{v}_k \|^2 \) is the squared Euclidean distance between embeddings \( \mathbf{v}_i \) and \( \mathbf{v}_k \).
\end{itemize}

\noindent
A higher value of \( d_i \) indicates that token \( i \) resides in a sparser region of the embedding space, thus exhibiting higher diversity. Based on this metric, the diversity penalty applied to a token pair \( (i, j) \) is defined as:
\[
w_{ij}^{\text{div}} = \exp\!\left(-\lambda_d (d_i + d_j)\right),
\]
where \( \lambda_d > 0 \) is a coefficient controlling the influence of the diversity term. This formulation downweights redundant token pairs in dense regions while amplifying the contribution of informative tokens during the matching process.

\paragraph{\textbf{Attention Preservation}}
The attention preservation component aims to precisely retain high-importance tokens during the pruning process, effectively preventing the undesired merging of critical visual and contextual information. This mechanism dynamically identifies and prioritizes key tokens based on their attention-derived importance scores, ensuring their preservation throughout the entire pruning pipeline.

For each token \( i \), its importance score \( s_i \) is computed by averaging the attention weights across all tokens and attention heads. Let \( \mathbf{A}^h \in \mathbb{R}^{N \times N} \) denote the attention matrix of the \( h \)-th head, where \( N \) is the number of tokens. The importance score is formulated as:
\[
s_i = \frac{1}{H} \sum_{h=1}^{H} \frac{1}{N} \sum_{j=1}^{N} A_{ij}^h,
\]
\noindent
where \( H \) is the number of attention heads and \( A_{ij}^h \) represents the attention weight from token \( i \) to token \( j \) in the \( h \)-th head. A higher \( s_i \) indicates that token \( i \) receives strong attention from other tokens and thus carries significant information.

To determine which tokens should be preserved, we define the \textbf{Critical Token Ratio (CTR)} that controls the proportion of tokens to protect. The number of protected tokens \( N_{\text{protect}} \) is given by:
\[
N_{\text{protect}} = \min(\text{CTR} \times T,\, T - 2r),
\]
\noindent
where \( T \) is the total number of tokens in the current layer and \( r \) is the number of tokens merged in each pruning step. This adaptive formulation ensures that the number of protected tokens decreases as pruning proceeds, avoiding excessive protection when \( T \) becomes small in later stages.

After computing \( s_i \), the top \( N_{\text{protect}} \) tokens are selected to form the critical token set \( \mathcal{S} \). These tokens are preserved in subsequent pruning stages to maintain essential information.

Finally, the \textbf{Attention Preservation} weight for each token pair \( (i, j) \) is defined as:
\[
w_{ij}^{\text{attn}} =
\begin{cases} 
\displaystyle -\infty, & \text{if } i \in \mathcal{S} \text{ or } j \in \mathcal{S}, \\[6pt]
\displaystyle 1, & \text{otherwise.}
\end{cases}
\]

Assigning \( w_{ij}^{\text{attn}} = -\infty \) to edges involving protected tokens prevents their merging, thus ensuring that high-importance tokens remain intact throughout the pruning process.

\section{Experiment}
\subsection{Experimental Settings}
\paragraph{\textbf{Model Architectures.}}
To ensure fair and reproducible comparisons, we follow established experimental setups in prior work on multimodal efficiency and token pruning. Image-based evaluations are conducted under the standard LLaVA-1.5-7B \cite{liu2024improved} configuration, while video-based evaluations follow the LLaVA-Video-7B \cite{zhang2024llavanext-video} setting with the Qwen2 \cite{yang2024qwen2technicalreport} language backbone. These complementary settings allow us to evaluate pruning strategies consistently across both static and temporal multimodal scenarios.

\paragraph{\textbf{Benchmarks.}}
To comprehensively evaluate both efficiency and generalization, we test our method on a diverse suite of vision-language benchmarks.For image understanding, we include VQAv2 \cite{jia2025vqa2}, MME \cite{fu2023mme}, MMBench \cite{mmbench} as MMB$^{EN}$ and the chinese version MMB$^{CN}$, and MMVet \cite{mmvet}, which collectively cover reasoning, perception, multilingual comprehension, and fine-grained visual recognition tasks.  
For video understanding, we evaluate on MVBench \cite{mvbench}, LongVideoBench \cite{LongVideoBench}, and Video-MME \cite{videomme}, each designed to assess different dimensions of temporal reasoning, motion perception, and long-term consistency. This combination of datasets ensures a fair and holistic evaluation of pruning performance across various multimodal contexts.

\paragraph{\textbf{Compared Methods.}}
We compare our method against a range of state-of-the-art visual token pruning approaches covering different methodological categories. Attention-based methods include \textit{FasterVLM} \cite{zhang2024cls}, which leverages attention scores for dynamic token selection. Attention and similarity-based methods are represented by \textit{VisPruner} \cite{zhang2025beyond}, which combines multi-head attention and feature similarity cues for token reduction. Similarity-based methods consist of \textit{DART} \cite{wen2025stop} and \textit{DivPrune} \cite{alvar2025divprune}, both relying purely on token-level feature similarity to guide merging or pruning. Query and similarity-based methods involve \textit{CDPruner} \cite{zhang2025attention}, which introduces task-driven query features to refine the similarity-based pruning process. This diverse selection of baselines enables a balanced and comprehensive comparison across distinct pruning paradigms, ensuring that the observed improvements are not confined to a specific design philosophy or task domain.

\paragraph{\textbf{Implementation Details.}}
All experiments are conducted on an NVIDIA H20 GPUs to ensure consistent and efficient inference across models. During the \textit{Attention Preservation} phase, we set the \textbf{Critical Token Ratio (CTR)} to \textbf{0.25}, meaning that the top 25\% of tokens with the highest attention importance are preserved before pruning.

For layer-wise token merging, we adopt the \textit{skip} strategy in our experiments, where token merging is applied at every other visual encoder layer. The overall merging budget is evenly distributed across all merging layers, with a fixed number of merged tokens $r$ per operation.

\begin{table*}[htbp!]
\centering
\caption{\textbf{Comparison of state-of-the-art methods on image-based benchmarks.} \textbf{Relative Latency} represents the normalized time breakdown of the pre-LLM, LLM, and Overall stages with respect to the Native(Unpruned). Here, \textbf{Acc.} denotes the average performance across all benchmarks, while \textbf{Rel.} indicates the mean relative accuracy (\%) compared to the native baseline.}
\resizebox{\textwidth}{!}{
\begin{tabular}{l!{\vrule width 1.0pt}ccccc!{\vrule width 0.9pt}ccc!{\vrule width 0.9pt}cc}
\toprule
\rowcolor{gray!5}
\textbf{Method} & \multicolumn{5}{c!{\vrule width 1.1pt}}{\textbf{Benchmarks $\uparrow$}} & \multicolumn{3}{c!{\vrule width 1.0pt}}{\textbf{Relative Latency $\downarrow$}} & \multicolumn{2}{c}{\textbf{Avg. $\uparrow$}} \\
\rowcolor{gray!5}
\textbf{Metric} & \textbf{VQA$^{V2}$} & \textbf{MME} & \textbf{MMB$^{EN}$} & \textbf{MMB$^{CN}$} & \textbf{MMVet} & \textbf{pre-LLM} & \textbf{LLM} & \textbf{Overall} & \textbf{Acc.} & \textbf{Rel.} \\
\rowcolor{gray!50}
\multicolumn{11}{c}{\textit{Upper Bound, All 576 Tokens (100\%)}} \\
\rowcolor{gray!20}
Native & 78.5 & 1508 & 61.8 & 58.1 & 32.2 & 1.00 & 1.00 & 1.00 & 76.5 & 100.0\% \\
\rowcolor{gray!50}
\multicolumn{11}{c}{\textit{Retain 128 Tokens ($\downarrow$ 77.8\%)}} \\
\rowcolor{gray!5}
FasterVLM & 75.5 & 1353 & 58.3 & 54.8 & 33.1 & 1.15 & 0.77 & 0.85 & 72.3 & 94.6\% \\
\rowcolor{gray!5}
VisPruner & 75.8 & 1405 & 58.8 & 55.2 & 32.8 & 1.32 & 0.77 & 0.90 & 73.2 & 95.7\% \\
\rowcolor{gray!5}
DART & 74.7 & 1344 & 57.5 & 52.9 & 31.3 & 1.20 & 0.77 & 0.88 & 70.9 & 92.7\% \\
\rowcolor{gray!5}
DivPrune & 76.0 & 1400 & 59.3 & 54.8 & 31.5 & 1.16 & 0.77 & 0.85 & 72.9 & 95.3\% \\
\rowcolor{gray!5}
CDPruner & 76.6 & 1433 & 60.6 & 55.0 & 31.7 & 1.41 & 0.77 & 0.91 & \underline{73.9} & \underline{96.6\%} \\
\rowcolor{blue!10}
\textbf{EvoPrune (Ours)} & 76.8 & 1441 & 59.7 & 57.3 & 33.7 & \textbf{1.10} & \textbf{0.77} & \textbf{0.84} & \textbf{74.9} & \textbf{97.9\%} \\
\rowcolor{gray!50}
\multicolumn{11}{c}{\textit{Retain 64 Tokens ($\downarrow$ 88.9\%)}} \\
\rowcolor{gray!5}
FasterVLM & 72.7 & 1294 & 56.7 & 51.9 & 26.8 & 1.09 & 0.73 & 0.79 & 68.2 & 89.2\% \\
\rowcolor{gray!5}
VisPruner & 73.9 & 1321 & 57.7 & 52.2 & 29.5 & 1.35 & 0.73 & 0.86 & 69.9 & 91.3\% \\
\rowcolor{gray!5}
DART & 71.3 & 1250 & 53.1 & 49.1 & 28.8 & 1.13 & 0.73 & 0.80 & 66.2 & 86.6\% \\
\rowcolor{gray!5}
DivPrune & 74.1 & 1342 & 58.1 & 52.4 & 29.9 & 1.12 & 0.73 & 0.79 & 70.4 & 92.0\% \\
\rowcolor{gray!5}
CDPruner & 75.4 & 1418 & 59.1 & 53.3 & 30.3 & 1.38 & 0.73 & 0.86 & \underline{72.2} & \underline{94.4\%} \\
\rowcolor{blue!10}
\textbf{EvoPrune (Ours)} & 75.6 & 1414 & 58.3 & 54.6 & 32.4 & \textbf{1.07} & \textbf{0.73} & \textbf{0.77} & \textbf{72.9} & \textbf{95.3\%} \\
\rowcolor{gray!50}
\multicolumn{11}{c}{\textit{Retain 32 Tokens ($\downarrow$ 94.4\%)}} \\
\rowcolor{gray!5}
FasterVLM & 70.9 & 1190 & 53.7 & 47.6 & 26.5 & 1.09 & 0.71 & 0.77 & 64.5 & 84.4\% \\
\rowcolor{gray!5}
VisPruner & 72.1 & 1196 & 53.9 & 48.5 & 26.6 & 1.35 & 0.71 & 0.84 & 65.2 & 85.2\% \\
\rowcolor{gray!5}
DART & 67.1 & 1119 & 49.1 & 43.3 & 23.7 & 1.09 & 0.71 & 0.76 & 59.8 & 78.1\% \\
\rowcolor{gray!5}
DivPrune & 71.2 & 1287 & 55.7 & 49.1 & 26.9 & 1.10 & 0.71 & 0.77 & 66.8 & 87.4\% \\
\rowcolor{gray!5}
CDPruner & 73.6 & 1384 & 57.4 & 49.6 & 28.6 & 1.22 & 0.71 & 0.80 & \textbf{69.6} & \textbf{91.0\%} \\
\rowcolor{blue!10}
\textbf{EvoPrune (Ours)} & 73.1 & 1337 & 56.8 & 51.3 & 26.6 & \textbf{1.06} & \textbf{0.71} & \textbf{0.75} & \underline{68.7} & \underline{89.8\%} \\
\bottomrule
\end{tabular}}
\label{tab:image_results}
\end{table*}

\subsection{Main Results}
\label{sec:main_results}
\paragraph{\textbf{EvoPrune for image understanding}}
We evaluate EvoPrune on a suite of image-based benchmarks using LLaVA-1.5-7B as the underlying vision-language model.
We report latency broken down into its computational components 
(\textit{pre-LLM} and \textit{LLM}) together with the overall end-to-end latency, where \textit{pre-LLM} includes visual encoding and token pruning, \textit{LLM} refers to the language model inference time, and \textit{overall} denotes the end-to-end latency. 

As shown in Table~\ref{tab:image_results}, EvoPrune consistently achieves the best trade-off between efficiency and accuracy across token retention levels. When retaining 128 tokens—representing a 77.8\% reduction in visual tokens—EvoPrune attains the highest average accuracy (74.9) and relative performance (97.9\%) while yielding the lowest overall latency (0.84$\times$). Compared to the strongest competitor, CDPruner, EvoPrune improves average accuracy by \textbf{1.0} percentage points and reduces latency by \textbf{7.7\%}, highlighting its superior pruning effectiveness. Even under more aggressive compression (64 and 32 tokens), EvoPrune maintains strong performance while achieving the fastest inference time across all reported latency metrics. These results demonstrate that EvoPrune not only preserves the semantic richness in highly compressed representations but also generalizes effectively across diverse image understanding tasks, validating EvoPrune as a scalable and computation-efficient pruning framework for MLLMs.

\paragraph{\textbf{EvoPrune for video understanding}}
\begin{table*}[htbp!]
\centering
\caption{\textbf{Comparison of state-of-the-art methods on video-based benchmarks.} \textbf{Relative Latency} represents the normalized time breakdown of the pre-LLM, LLM, and Overall stages with respect to the Native(Unpruned). Here, \textbf{Acc.} denotes the average performance across all benchmarks, while \textbf{Rel.} indicates the mean relative accuracy (\%) compared to the native baseline.}
\resizebox{\textwidth}{!}{
\begin{tabular}{l!{\vrule width 1.0pt}cccccccc!{\vrule width 0.9pt}ccc!{\vrule width 0.9pt}cc}
\toprule
\rowcolor{gray!5}\textbf{Method} & \textbf{MVBench $\uparrow$} & \multicolumn{3}{c}{\textbf{LongVideoBench $\uparrow$}} & \multicolumn{4}{c}{\textbf{Video-MME $\uparrow$}} & \multicolumn{3}{c}{\textbf{Relative Latency $\downarrow$}} & \multicolumn{2}{c}{\textbf{Avg. $\uparrow$}} \\ 
\rowcolor{gray!5}
\textbf{Metric} & \textbf{test} & \textbf{val} & perception & relation & \textbf{w/o sub} & short & medium & long & \textbf{pre-LLM} & \textbf{LLM} & \textbf{Overall} & \textbf{Acc.} & \textbf{Rel.} \\ 
\rowcolor{gray!50}
\multicolumn{14}{c}{\textit{Upper Bound, All 64 $\times$ 169 Tokens (100\%)}} \\
\rowcolor{gray!20}
Native & 58.0 & 56.0 & 64.0 & 48.9 & 59.7 & 71.3 & 58.0 & 49.7 & 1.00 & 1.00 & 1.00 & 57.9 & 100.0\% \\
\rowcolor{gray!50}
\multicolumn{14}{c}{\textit{Retain 64 $\times$ 64 Tokens ($\downarrow$ 62.1\%)}} \\
\rowcolor{gray!5}
FasterVLM & 58.3 & 54.9 & 61.6 & 49.0 & 59.4 & 70.1 & 57.4 & 50.6 & 1.01 & 0.45 & 0.77 & 57.5 & 99.4\% \\
\rowcolor{gray!5}
VisPruner & 58.3 & 54.9 & 61.6 & 49.0 & 59.4 & 70.1 & 57.4 & 50.6 & 1.01 & 0.45 & 0.77 & 57.5 & 99.4\% \\
\rowcolor{gray!5}
DART & 57.8 & 54.2 & 61.3 & 47.9 & 59.3 & 70.6 & 57.3 & 50.1 & 1.04 & 0.45 & 0.78 & 57.1 & 98.7\% \\
\rowcolor{gray!5}
DivPrune & 57.8 & 54.8 & 62.4 & 48.2 & 60.0 & 71.2 & 58.0 & 50.8 & 1.01 & 0.45 & 0.77 & 57.5 & 99.4\% \\
\rowcolor{gray!5}
CDPruner & 58.1 & 55.1 & 61.3 & 49.7 & 59.6 & 70.8 & 57.0 & 51.0 & 1.01 & 0.45 & 0.77 & \underline{57.6} & \underline{99.5\%} \\
\rowcolor{blue!10}
\textbf{EvoPrune (Ours)} & 56.9 & 55.5 & 61.9 & 49.7 & 60.6 & 70.9 & 58.9 & 51.9 & \textbf{0.54} & \textbf{0.45} & \textbf{0.55} & \textbf{57.7} & \textbf{99.7\%} \\
\rowcolor{gray!50}
\multicolumn{14}{c}{\textit{Retain 64 $\times$ 32 Tokens ($\downarrow$ 81.1\%)}} \\
FasterVLM & 56.8 & 52.7 & 58.9 & 47.2 & 58.1 & 67.1 & 56.6 & 50.6 & 1.01 & 0.30 & 0.71 & 55.8 & 96.5\% \\
\rowcolor{gray!5}
VisPruner & 57.2 & 53.5 & 60.0 & 47.8 & 58.3 & 69.4 & 56.2 & 49.2 & 1.01 & 0.30 & 0.71 & 56.3 & 97.3\% \\
\rowcolor{gray!5}
DART & 57.3 & 53.5 & 60.8 & 47.2 & 59.0 & 69.3 & 57.7 & 50.3 & 1.02 & 0.30 & 0.71 & 56.6 & 97.8\% \\
\rowcolor{gray!5}
DivPrune & 56.9 & 54.4 & 60.8 & 48.0 & 58.6 & 68.4 & 56.9 & 50.5 & 1.01 & 0.30 & 0.71 & 56.6 & 97.8\% \\
\rowcolor{gray!5}
CDPruner & 57.2 & 54.5 & 61.8 & 48.0 & 58.6 & 68.8 & 56.3 & 50.8 & 1.01 & 0.30 & 0.71 & \underline{56.7} & \underline{97.9\%} \\
\rowcolor{blue!10}
\textbf{EvoPrune (Ours)} & 56.1 & 55.1 & 62.7 & 48.5 & 58.7 & 68.3 & 57.2 & 50.6 & \textbf{0.50} & \textbf{0.30} & \textbf{0.47} & \textbf{56.7} & \textbf{98.0\%} \\
\rowcolor{gray!50}
\multicolumn{14}{c}{\textit{Retain 64 $\times$ 16 Tokens ($\downarrow$ 90.5\%)}} \\
\rowcolor{gray!5}
FasterVLM & 54.5 & 51.2 & 57.0 & 46.1 & 55.8 & 63.8 & 54.1 & 49.6 & 1.01 & 0.24 & 0.68 & 53.8 & 93.0\% \\
\rowcolor{gray!5}
VisPruner & 55.2 & 51.4 & 58.9 & 44.8 & 56.3 & 65.9 & 54.0 & 49.1 & 1.01 & 0.24 & 0.68 & 54.3 & 93.8\% \\
\rowcolor{gray!5}
DART & 55.5 & 53.0 & 61.1 & 45.9 & 57.6 & 67.6 & 56.0 & 49.3 & 1.02 & 0.24 & 0.68 & \underline{55.4} & \underline{95.7\%} \\
\rowcolor{gray!5}
DivPrune & 56.0 & 53.1 & 60.3 & 46.8 & 56.9 & 66.1 & 55.3 & 49.2 & 1.01 & 0.24 & 0.68 & 55.3 & 95.6\% \\
\rowcolor{gray!5}
CDPruner & 55.9 & 52.7 & 60.5 & 45.8 & 57.2 & 66.7 & 55.4 & 49.6 & 1.01 & 0.24 & 0.68 & 55.3 & 95.5\% \\
\rowcolor{blue!10}
\textbf{EvoPrune (Ours)} & 55.8 & 53.2 & 58.9 & 48.2 & 57.4 & 65.8 & 56.0 & 50.3 & \textbf{0.46} & \textbf{0.24} & \textbf{0.43} & \textbf{55.4} & \textbf{95.8\%} \\
\bottomrule
\end{tabular}}
\label{tab:video_results}
\end{table*}
In the video domain, the benefits of EvoPrune's \textbf{Early-Stage Visual Token Pruning} are particularly pronounced. For video-based evaluation, we employ LLaVA-Video-7B \cite{zhang2024llavanext-video} as the backbone model. All videos are uniformly sampled to 64 frames and then resized to a resolution of $384 \times 384$. Following the intermediate pooling operations within LLaVA-Video, each video comprises $64 \times 169$ visual tokens prior to pruning. 

Notably, the pooling occurs between the visual encoder and the LLM, meaning that all baseline methods perform token pruning on an already pooled—and thus significantly reduced—set of tokens. In contrast, EvoPrune conducts pruning earlier within the visual encoder, requiring the handling of a substantially larger token set. This early-stage pruning presents a greater reliability challenge, as the method must identify and retain the most informative tokens before pooling while preserving overall model performance.

Table~\ref{tab:video_results} provides a detailed comparison across multiple video benchmarks. When retaining $64 \times 64$ tokens (62.1\% reduction), EvoPrune achieves an average accuracy of 57.7, surpassing all baselines while reducing relative overall latency to 0.55$\times$.

At a more aggressive pruning level of $64 \times 32$ tokens (81.1\% reduction), EvoPrune maintains an relative accuracy of 98.0\%, outperforming most baselines and reducing both pre-LLM and overall latency by nearly half (0.50$\times$ and 0.47$\times$, respectively). Even under extreme pruning with $64 \times 16$ tokens (90.5\% reduction), EvoPrune preserves 95.8\% relative accuracy (55.4 avg.) and lowers overall latency to 0.43$\times$ vs. 0.68$\times$ of others, demonstrating robustness under heavy token reduction. In contrast, most baselines suffer larger accuracy drops or achieve smaller latency reductions, highlighting EvoPrune's favorable efficiency-effectiveness trade-off.

Overall, these results demonstrate EvoPrune can reliably perform early-stage token pruning within the visual encoder, substantially lowering computational cost while maintaining high performance, even when operating on a considerably larger set of tokens compared to baseline methods.

\subsection{Efficiency Results}
\begin{figure*}[htbp!]
  \centering
  \includegraphics[width=0.9\textwidth]{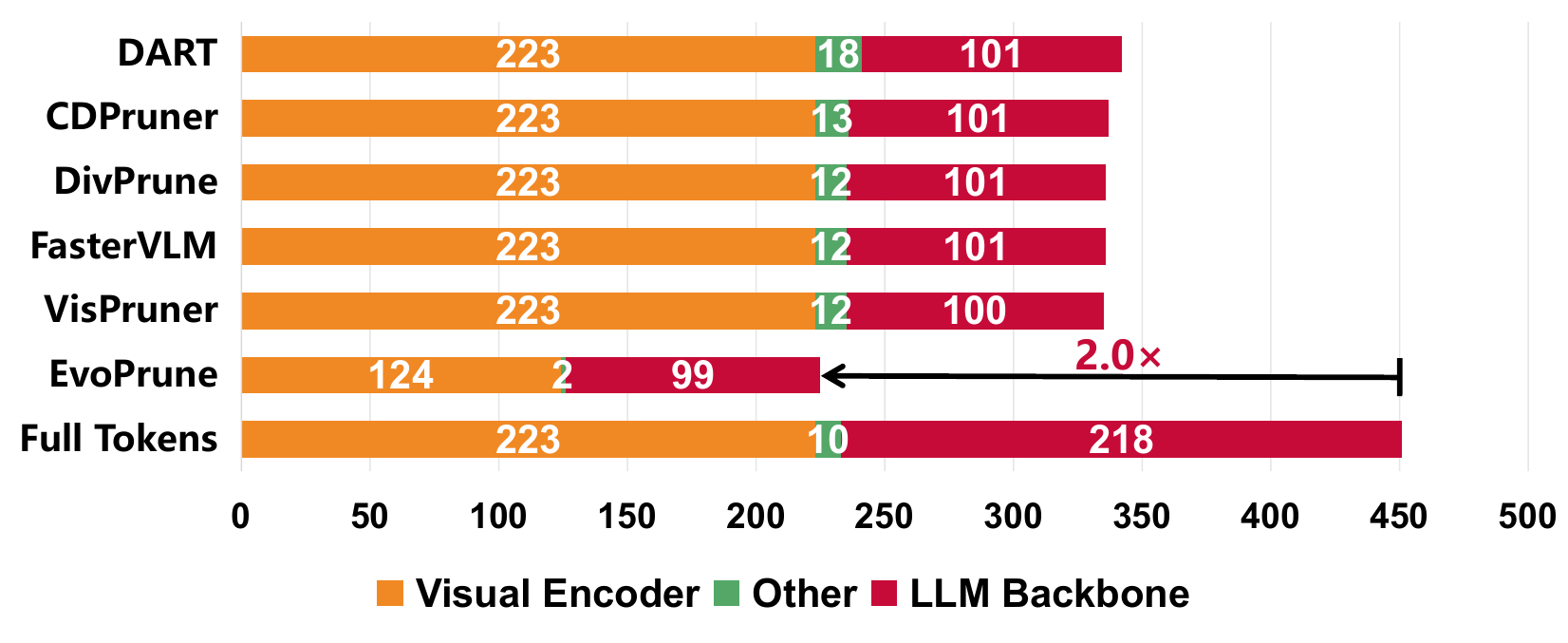}
  \caption{\textbf{Breakdown of Time-To-First-Token (TTFT, 1 unit = 10 ms).} The \textcolor{orange}{Visual  Encoder}, \textcolor{green}{Other}, and \textcolor{red}{LLM Backbone} correspond to the visual encoder, intermediate processing modules (e.g., pooling), and the language model backbone, respectively.
  }
  \label{fig:ttft_breakdown}
\end{figure*}

We evaluate the efficiency of EvoPrune against SOTA visual token pruning methods on the Video-MME dataset. Figure~\ref{fig:ttft_breakdown} presents a breakdown of  \textbf{Time-To-First-Token (TTFT)} across three stages: the visual encoder, intermediate modules (e.g., pooling), and the LLM backbone.  

Existing methods—including VisPruner, FasterVLM, DivPrune, CDPruner, and DART—exhibit clear bottlenecks in pruning efficiency, particularly in the video domain. While these approaches can reduce computation within the LLM backbone, the visual encoder remains largely unoptimized. Consequently, they achieve comparable speedups when pruning the same number of tokens, resulting in limited overall acceleration (1.3$\times$-1.4$\times$), as the visual encoder dominates the computational cost.

In contrast, EvoPrune performs early-stage token pruning within the visual encoder, yielding a 1.8$\times$ speedup at this stage. It further accelerates intermediate processing by 5$\times$ and reduces LLM backbone computation by approximately 2$\times$. Overall, EvoPrune achieves a \textbf{2$\times$} reduction in TTFT compared to the full-token baseline, demonstrating consistent end-to-end acceleration across all pipeline stages rather than being confined to the LLM.

These results demonstrate that EvoPrune preserves accuracy (Section~\ref{sec:main_results}) while achieving substantial practical speedups. By jointly accelerating the visual encoder, intermediate modules, and the LLM backbone, EvoPrune is well-suited for real-time and large-scale video understanding.

\subsection{Ablation Study.}
\begin{table*}[htbp!]
\centering
\small
\resizebox{0.8\textwidth}{!}{
\begin{tabular}{l!{\vrule width 0.8pt}ccc!{\vrule width 1.0pt}cc}
\toprule
\rowcolor{gray!5}
\textbf{Method} & \textbf{MVBench} & \textbf{LongVideoBench} & \textbf{VideoMME} & \textbf{Avg.} & \textbf{Rel.} \\
\rowcolor{gray!20}
Native & 58.0 & 56.0 & 59.7 & 57.9 & 100.0\% \\
\rowcolor{gray!5}
EvoPrune (w/o Attn\&Div) & 54.6 & 54.3 & 59.0 & 55.9 & 96.5\% \\
\rowcolor{gray!5}
EvoPrune (w/o Attn) & 55.5 & 54.9 & 59.7 & 56.7 & 97.9\% \\
\rowcolor{gray!5}
EvoPrune (w/o Div) & 56.2 & 55.2 & 60.4 & 57.3 & 98.9\% \\
\rowcolor{gray!5}
\textbf{EvoPrune} & \textbf{56.9} & \textbf{55.5} & \textbf{60.6} & \textbf{57.7} & \textbf{99.7\%} \\
\bottomrule
\end{tabular}}
\caption{\textbf{Ablation study on score factors.} We evaluate the contribution of each component—Attention Preservation and Diversity Penalty—by disabling them individually. Results are reported on MVBench, LongVideoBench, and VideoMME. \textbf{Avg.} denotes the mean score across benchmarks, and \textbf{Rel.} is the accuracy relative to the unpruned model.}
\label{tab:evoprune_ablation}
\end{table*}

To analyze the contribution of each scoring component in \textbf{EvoPrune}, we conduct an ablation study on three video understanding benchmarks, with results summarized in Table~\ref{tab:evoprune_ablation}. We evaluate variants that remove the \textit{Attention Preservation} term, the \textit{Diversity Penalty}, or both.

Removing both components (w/o Attn\&Div) results in the largest performance drop, reducing the average score from \textbf{57.7} to \textbf{55.9}, indicating naive pruning without explicit saliency or diversity constraints is insufficient for long video understanding. When disabling individual components, removing attention preservation (w/o Attn) leads to a larger degradation than removing the diversity penalty (w/o Div), suggesting that attention-critical tokens play a more prominent role in maintaining spatiotemporal semantics.

The full EvoPrune model achieves the best performance, with an average score of \textbf{57.7} and \textbf{99.7\%} relative accuracy. These results demonstrate that attention-guided preservation and diversity-aware selection contribute complementary benefits, and their combination is essential for robust and efficient token pruning in video understanding tasks.

\section{Conculsion}
We propose EvoPrune, a plug-and-play visual token pruning method for MLLMs that addresses inference inefficiency from redundant visual tokens. By enabling early-stage pruning within the visual encoder and layer-wise token fusion, EvoPrune achieves end-to-end acceleration while preserving critical visual information at high pruning rates. The method requires no retraining and can be seamlessly integrated into existing MLLM architectures. Extensive experiments on diverse benchmarks demonstrate that EvoPrune consistently outperforms existing pruning approaches in terms of the speed--accuracy trade-off, particularly in scenarios with high-resolution images and dense visual inputs.

Future work will extend EvoPrune to more complex video settings, such as long and dynamic sequences. In particular, incorporating temporal-aware pruning strategies that leverage cross-frame redundancy may further improve efficiency and scalability for real-world multimodal applications.


\bibliography{references}  

\clearpage
\appendix
\section{Analysis of Layer-wise Merging Strategies}
\label{sec:appendix_merging_strategy}
This appendix provides a detailed analysis of the layer-wise token merging strategies explored in EvoPrune. Our goal is to understand how different distributions of the pruning budget across encoder layers affect the trade-off between inference efficiency and model performance.

\begin{figure*}[htbp!]
  \centering
  \includegraphics[width=1.0\textwidth]{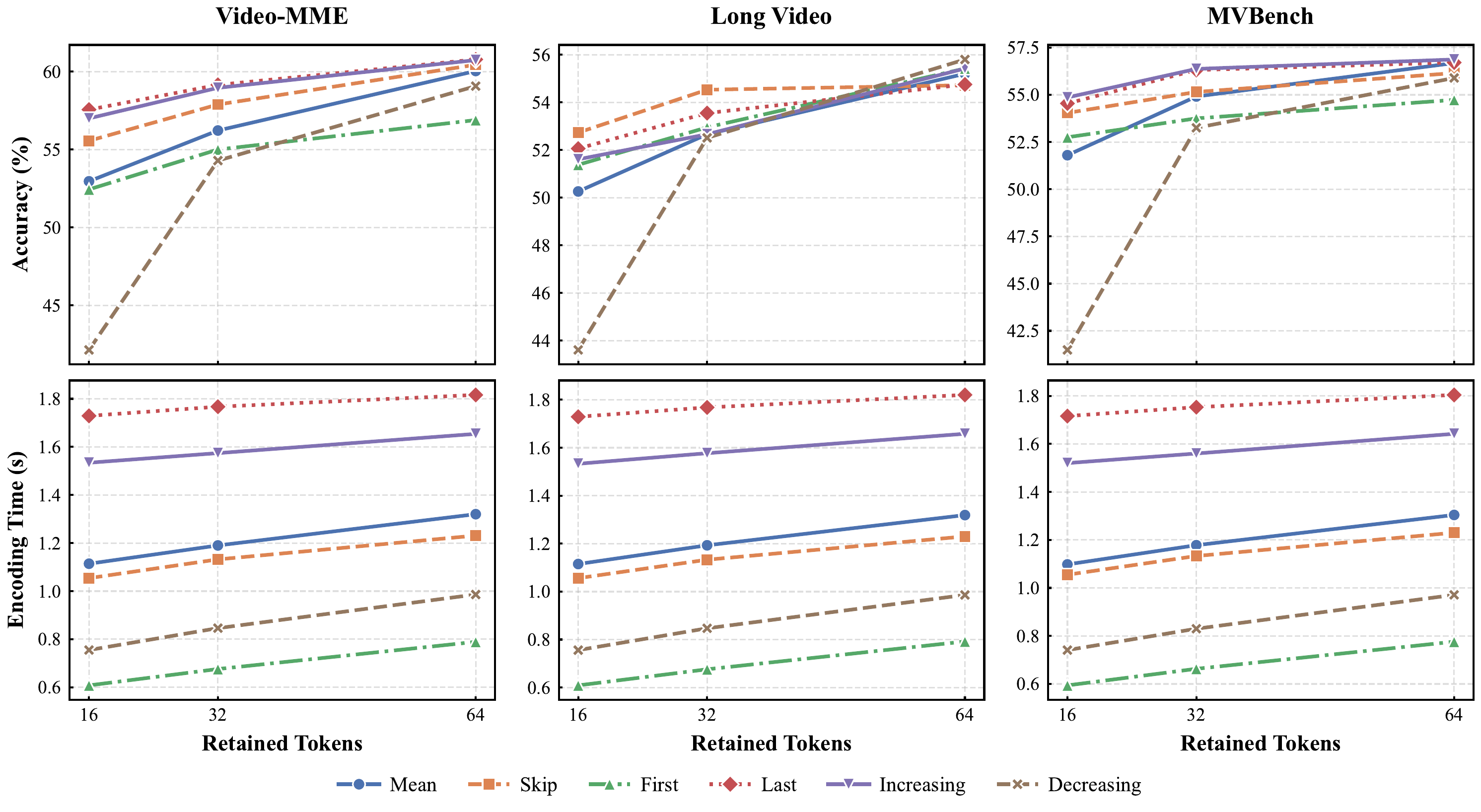}
  \caption{
    \textbf{Performance comparison of different layer-wise merging strategies across video benchmarks}
    We present the Accuracy (top row) and Encoding Time (bottom row) as functions of the retained token budget $B \in \{16, 32, 64\}$.
    Each curve represents a specific layer-wise allocation pattern. Specifically, we set the window size $N=13$ for \textit{First} and \textit{Last} (out of 26 total layers), and the growth/decay rate $\alpha=1$ for \textit{Increasing} and \textit{Decreasing}.
    }
  \label{fig:merge_strategy_comparison}
\end{figure*}

\subsection{Layer-wise Merging Strategy Design}
We explore a set of layer-wise token merging strategies to investigate how different distributions of the pruning budget $\mathbf{r}$ across encoder layers affect model efficiency and performance. Each strategy specifies how the total pruning budget $R$ is allocated among selected layers.

\begin{itemize}
  \item \textbf{Mean:} The pruning budget is uniformly distributed across all selected encoder layers, assigning equal weights to each layer.

  \item \textbf{Skip:} Token merging is applied at alternating layers, where only every other layer participates in pruning.

  \item \textbf{First[$N$]:} Token merging is restricted to the first $N$ encoder layers, while the remaining layers are left unpruned.

  \item \textbf{Last[$N$]:} Token merging is applied only to the last $N$ encoder layers, concentrating the pruning budget on deeper representations.

  \item \textbf{Increasing[$\alpha$]:} The pruning budget increases with layer depth, assigning progressively larger budgets to deeper layers. The parameter $\alpha$ controls the growth rate, with $\alpha = 1$ by default.

  \item \textbf{Decreasing[$\alpha$]:} The pruning budget decreases with layer depth, emphasizing stronger pruning in earlier layers. The parameter $\alpha$ similarly controls the decay rate, with $\alpha = 1$ by default.
\end{itemize}

All strategies are normalized to ensure that the total pruning budget equals $R$. This design space enables flexible control over where pruning is applied during visual encoding, allowing us to study the trade-offs between early-stage acceleration and representational capacity under different layer-wise pruning patterns.

\subsection{Experimental Results and Analysis}

We evaluate the above strategies on representative image and video benchmarks under the same overall pruning budget. The results reveal clear differences in efficiency--accuracy trade-offs among the strategies.
For clarity, we denote each experimental setting as \textit{merging strategy + token budget}. For example, \textit{First16} indicates the \textit{First} merging strategy with a token budget of 16.

Our results indicate that the stage at which token pruning is applied significantly influences both performance and inference speed. 
Strategies that concentrate pruning at the early stages, such as \textit{First} and \textit{Decreasing}, exhibit the lowest inference times (e.g., \textit{First16} reaches 0.608s on Video-MME). 
However, they suffer from a drastic degradation in accuracy, particularly in \textit{Decreasing16} where the accuracy on MVBench drops to 41.49\%. 
This suggests that aggressive pruning in early layers destroys fine-grained spatial-temporal features before they can be effectively integrated. 
Conversely, \textit{Last} and \textit{Increasing} strategies maintain high accuracy (up to 60.78\% on Video-MME) by preserving full token density in earlier layers, but they are computationally expensive, with \textit{Last64} being nearly 1.5$\times$ slower than the average.

Among all configurations, the \textbf{Skip} strategy provides the most robust Pareto frontier. 
By applying token merging at alternating layers, it prevents the accumulation of pruning error in any single stage of the visual encoder. 
Quantitatively, \textit{Skip64} achieves a superior accuracy of 60.44\% on Video-MME with a latency of 1.231s, outperforming the \textit{Mean} strategy (60.04\%, 1.320s) in both metrics. 
Even at the most aggressive budget ($B=16$), \textit{Skip} maintains 55.56\% accuracy on Video-MME, significantly higher than \textit{Mean} (52.96\%) and \textit{First} (52.44\%). 
This demonstrates that a staggered pruning pattern effectively balances the preservation of representational capacity with computational efficiency. 
Consequently, we adopt the \textit{Skip} strategy as our default allocation pattern for all experiments.

\section{Benchmarks}
\subsection{Image benchmarks}
\paragraph{VQAv2 \cite{jia2025vqa2}}
A large-scale Visual Question Answering benchmark that emphasizes true visual reasoning by mitigating language priors inherent in earlier VQA datasets. Built from images in MS-COCO with paired natural language questions and 10 human-annotated answers, VQAv2 collects “complementary” image pairs where the same question yields different answers, forcing models to rely on visual content rather than statistical shortcuts. It contains ~1.1M image-question pairs across train/val/test and is widely used to evaluate open-ended VQA performance using standard accuracy metrics.

\paragraph{MME \cite{fu2023mme}}
A comprehensive evaluation benchmark for Multimodal Large Language Models (MLLMs), designed to jointly assess both perception and high-level cognition across 14 diverse subtasks. To ensure fair comparison and avoid leakage from public datasets, instruction-answer pairs are manually crafted with concise prompts, minimizing prompt engineering bias. MME provides quantitative metrics and a systematic leaderboard of 30 advanced MLLMs, revealing persistent gaps in current multimodal understanding and guiding future model improvements.

\paragraph{MMBench \cite{mmbench}}
A finely-designed evaluation suite that systematically assesses the multimodal perception and reasoning skills of vision-language models across diverse visual understanding tasks. It comprises a large, meticulously curated set of bilingual multiple-choice questions spanning dozens of fine-grained abilities, with rigorous quality control and a novel CircularEval strategy to convert free-form model outputs into standardized choices. Integrated into VLMEvalKit, MMBench offers robust, scalable, and objective benchmarking that reveals distinct capability profiles of contemporary multimodal models.

\paragraph{MMVet \cite{mmvet}}
A evaluation benchmark for large multimodal models, designed to assess their integrated vision-language reasoning beyond isolated tasks. It defines six core vision-language capabilities—recognition, knowledge, spatial awareness, language generation, OCR, and mathematics—and examines their 16 pairwise integrations through open-ended image-question pairs. A unified LLM-based evaluator scores free-form responses, enabling consistent comparison across diverse question types and answer styles. MM-Vet thus provides insight into models’ holistic multimodal competence, revealing strengths and weaknesses beyond aggregate scores.

\subsection{Video benchmarks}
\paragraph{MVBench \cite{mvbench}}
A benchmark emphasizing dynamic temporal reasoning beyond static frame analysis to rigorously assess multimodal large language models’ video understanding capabilities. It defines 20 temporally-oriented tasks by systematically transforming traditional spatial vision tasks into sequence-dependent challenges, with each task instantiated as multiple-choice QA derived from public video annotations over diverse sources. This design enables fair, scalable evaluation of perception and cognition across actions, interactions, motion, and event reasoning, revealing substantial gaps in current model performance. The metric of interest is task accuracy over the full suite of temporal tests.

\paragraph{LongVideoBench \cite{LongVideoBench}}
A long-context question-answering benchmark that probes the ability of multimodal models to comprehend interleaved video and language inputs spanning up to an hour, requiring detailed retrieval and reasoning over extended temporal sequences rather than isolated frames. It comprises 3,763 web-sourced videos and 6,678 human-annotated multiple-choice questions across 17 fine-grained referring-reasoning categories, explicitly designed to assess long-form multimodal understanding. Evaluations show significant challenges for current LMMs, with performance gains tied to processing denser frame inputs, positioning it as a rigorous testbed for future long-context LMMs.

\paragraph{Video-MME \cite{videomme}}
A full-spectrum evaluation benchmark for multimodal large language models on video analysis, designed to stress temporal, domain, and multimodal comprehension beyond static frames. It comprises 900 manually curated videos totaling 254 hours across six diverse visual domains and 30 subcategories, with 2,700 expert-annotated multiple-choice QA pairs probing recognition, temporal reasoning, and event understanding using both frames and auxiliary modalities (subtitles, audio). Videos span short (<2min), medium (4–15min) and long (30–60min) durations, enabling rigorous assessment of long-context and multimodal capabilities. Standardized accuracy metrics facilitate comparison across state-of-the-art models.

\section{Compared Methods}
\paragraph{FasterVLM \cite{zhang2024cls}}
Existing visual token pruning in VLMs often relies on text-visual cross-attention, which misaligns with true token importance and degrades accuracy at high pruning rates. FasterVLM proposes a training-free method that uses [CLS] attention from the visual encoder to more accurately rank and prune visual tokens before they enter the multimodal projection, reducing redundant tokens without LLM interaction. This early pruning significantly speeds up inference while retaining the majority of performance across various VLM architectures and reduction ratios.

\paragraph{VisPruner \cite{zhang2025beyond}}
The first work to demonstrate that text-visual attention scores are a suboptimal criterion for visual token pruning in large vision-language models. Instead, it exploits intrinsic visual cues to identify and retain a compact set of informative tokens and removes redundant ones via similarity filtering, preserving diverse visual information. This plug‐and‐play approach sustains performance across VLM architectures without any training, achieving substantial FLOPs and latency reduction while maintaining comparable accuracy to full models.

\paragraph{DART \cite{wen2025stop}}
The first framework to explicitly enforce disease‐relevant alignment between visual findings and textual content in radiology report generation. It formulates a two‐stage pipeline where disease‐aware image-text retrieval based on contrastive learning is used to obtain initial reports that closely match pathological patterns, followed by a self‐correcting re‐alignment module that refines the generated text for clinical consistency. Evaluated on MIMIC‐CXR and IU X‐ray benchmarks, DART achieves state‐of‐the‐art performance in both report quality and disease classification, demonstrating improved trustworthiness of automated radiology reports.

\paragraph{DivPrune \cite{alvar2025divprune}}
Formulates visual-token pruning as a Max-Min Diversity Problem to explicitly maximize representational diversity among retained tokens, rather than relying on attention scores or calibration. It greedily selects a subset of visual tokens with maximal pairwise distances, pruning the rest to reduce redundancy and preserve broad visual information without fine-tuning. DivPrune operates as a plug-and-play module for large multimodal models, achieving state-of-the-art accuracy across 16 image/video language benchmarks while significantly reducing inference latency and memory usage.

\paragraph{CDPruner \cite{zhang2025attention}}
A novel training-free strategy that departs from traditional attention- or similarity-based pruning by maximizing the conditional diversity of retained visual tokens conditioned on the instruction. CDPruner defines conditional similarity between tokens and formulates pruning as a determinantal point process (DPP) to select a diverse subset that aligns with the user query. This model-agnostic method significantly reduces FLOPs and latency while preserving task accuracy across diverse vision-language benchmarks, establishing new state-of-the-art performance.

\end{document}